\documentclass[a4paper]{article}
\usepackage{iwslt14,amssymb,amsmath,epsfig}
\usepackage{url}
\usepackage{latexsym}
\usepackage{amsfonts}
\usepackage{color}
\usepackage{epstopdf}
\usepackage{multirow}
\usepackage[normalem]{ulem}

\def\reg{{\rm\ooalign{\hfil
     \raise.07ex\hbox{\scriptsize R}\hfil\crcr\mathhexbox20D}}}
     
\renewcommand{\vec}[1]{\mathbf{#1}}

\title{Lexical Translation Model Using a Deep Neural Network Architecture}
\name{Thanh-Le Ha, Jan Niehues, Alex Waibel}

\address{International Center for Advanced Communication Technologies - InterACT\\
Institute of Anthropomatics and Robotics \\
Karlsruhe Institute of Technology, Germany \\ 
{\small \tt  \{thanh-le.ha|jan.niehues|alex.waibel\}@kit.edu}}
\begin{document}
\maketitle

\begin{abstract}
In this paper we combine the advantages of a model using global source sentence contexts, the Discriminative Word Lexicon, and neural networks. 
By using deep neural networks instead of the linear maximum entropy model in the Discriminative Word Lexicon models, 
we are able to leverage dependencies between different source words due to the non-linearity.  
Furthermore, the models for different target words can share parameters and therefore data sparsity problems are effectively reduced.

By using this approach in a state-of-the-art translation system, we can improve the performance by up to 0.5 BLEU points for three different language pairs on the TED translation task.
\end{abstract}

\section{Introduction}
\label{sec:introduction}

Since the first attempt to statisical machine translation (SMT) \cite{ref:brown1993}, the approach has drawn much interest in the research community and
huge improvements in translation quality have been achieved. Still, there are plenty of problems in SMT which should be addressed. 
One is that the translation decision depends on a quite small context.

In standard phrase-based statistical machine translation (PBMT) \cite{ref:koehn2003}, the two main components are the translation and language models. 
The translation model is modeled by counting phrase pairs, which are sequences of words extracted from bilingual corpora. 
By using phrase segments instead of words, PBMT can exploit some local source and target contexts within those segments. 
But no context information outside the phrase pairs is used. In an $n$-gram language model, only a context of up to $n$ target words is considered.


Several directions have been proposed to leverage information from wider contexts in the phrase-based SMT framework.  
For example, the Discriminative Word Lexicon (DWL) \cite{ref:mauser2009}\cite{niehues2013mt} exploits the occurence of all the words in the whole source sentence to predict the presence of words in the target sentence. 
This wider context information is encoded as features and employed in a discriminative framework. 
Hence, they train a maximum entropy (MaxEnt) model for each target word.

While this model can improve the translation quality in different conditions, MaxEnt models are linear classifiers. 
On the other hand, hierarchical non-linear classifiers can model dependencies between different source words better since they perform some abstraction over the input. 
Hence, introducing non-linearity into the modeling of the lexical translation could improve the quality.
Moreover, since many pairs of source and target words co-occur only rarely, a way of sharing information between the different classifiers could improve the modeling as well.

In order to address these issues, we developed a discriminative lexical model based on deep neural networks. 
Since we train one neural network for all target words as a multivariate binary classifier, the model can share information between different target words.
Furthermore, the probability is no longer a linear combination of weights depending on the surface source words. 
Thanks to the non-linearity, we are now able to exploit semantic dependencies among source words.

This paper is organized as follows. 
In Section~\ref{sec:relatedwork}, we review the previous works related to lexical translation methods as well as the translation modeling using neural networks.
Then we describe our approach including the network architecture and its training procedures in Section~\ref{sec:NNDWL}.
Section~\ref{sec:experiments} provides experimental results of our translation systems for different language pairs using the proposed lexical translation model.
Finally, the conclusions are drawn in Section ~\ref{sec:conclusion}. 

\section{Related work}
\label{sec:relatedwork}
Since the beginnings of SMT, several approaches to increase the context used for lexical decisions have been presented. 
When moving from word-based to phrase-based SMT \cite{ref:koehn2003}\cite{och2004alignment}, a big step in employing wider contexts into translation systems has been made. 
In PBMT, the lexical joint models allow us to use local source and target contexts in the form of phrases.
Lately, advanced joint models have been proposed to either enhance the joint probability model between source and target sides or  engage more suitable contexts. 

The $n$-gram based approach \cite{marino2006} directly models the joint probability of source and target sentences 
from the conditional probability of a current $n$-gram pair givens sequences of previous bilingual $n$-grams. 
In \cite{niehues2011}, this idea is introduced into the phrase-based MT approach. Thereby, parallel context over phrase boundaries can be used during the translation.

Standard phrase-based or $n$-gram translation models are basically built upon statistical principles such as Maximum Entropy and smoothing techniques. 
Recently, joint models are learned using neural networks where non-linear translation relationships and semantic generalization of words can be performed \cite{mikolov2013linguistic}. 
Le et. al. \cite{son2012continuous} follow the $n$-gram translation direction 
but model the conditional probability of a target word given the history of bilingual phrase pairs using a neural network architecture. 
They then use their model in a $k$-best rescorer instead of in their $n$-gram decoder. 
Devlin et. al. \cite{devlin2014fast} add longer source contexts and renew the joint formula so that it can be included in a decoder rather than a $k$-best rescoring module.  
Schwenk et. al. \cite{schwenk2012continuous} calculate the conditional probability of a target phrase instead of a target word given a source phrase.    

Although the aforementioned works essentially augment the joint translation model, they have an inherent limitation: only exploit local contexts. 
They estimate the joint model using sequences of words as the basic unit. On the other hand, there are several approaches utilizing global contexts. 
Motivated by Bangalore et. al \cite{bangalore2007statistical}, Hasan et. al. \cite{hasan2008triplet} calculate the probability of a target word given two source words which do not necessarily belong to a phrase. 
Mauser et. al. \cite{ref:mauser2009} suggest another lexical translation approach, named Discriminative Word Lexicon (DWL), concentrating on predicting the presence of target words given the source words. 
Niehues et. al. \cite{niehues2013mt} extend the model to employ the source and target contexts, but they used the same MaxEnt classifier for the task. 
Carpuat et. al. \cite{carpuat2007improving} is the most similar work to the DWL direction in terms of using the whole source sentence to perform the lexical choices of target words. 
They treat the selection process as a Word Sense Disambiguation (WSD) task, where target words or phrases are WSD senses. 
They extract a rich feature set from the source sentences, including source words, and input them into a WSD classifier. 
Still, the problem persists since they use the shallow classifiers for that task.  

Considering the advantages of non-linear models mentioned before, we opt for using deep neural network architectures to learn the DWL.
We take the advantages of the two directions. On one side, our model uses a non-linear classification method to leverage dependencies between different source sentences 
as well as its semantic generalization ability. 
On the other side, by employing the global contexts, our model can complement joint translation models  which use the local contexts.

\section{Discriminative lexical translation using deep neural networks}
\label{sec:NNDWL}
We will first review the original DWL approach described in \cite{ref:mauser2009} and \cite{niehues2013mt}. 
Afterwards, we will describe the neural network architecture and training procedures proposed in this work. 
We will finish this section by describing the integration into the decoding process.

\subsection{Original Discriminative Word Lexicon}
\label{DWL}
In this approach, the DWL are modeled using a maximum entropy model to determine the probability of using a target word in the translation. 
Therefore, individual models for every target word are trained. Each model is trained to return the probability of this word given the input sentence.

The input of the model is the source sentence, thus, they need a way to represent the input sentence. 
This is done by representing the sentence as a bag of words and thereby ignoring the order of the words.
In the MaxEnt model, they use an indicator feature for every input word.
More formally, a given source sentence $s=s_1 \ldots s_I$ is represented by the features $F(s)=\{f_w(s): \forall{w \in V_s}\}$, with $V_s$ is the source vocabulary:
\begin{equation}
   f_w(s) =  \left\{
     \begin{array}{lr}
       1 & \text{if} \quad w \in s\\
       0 & \text{if} \quad w \notin s
     \end{array}
   \right.
\end{equation}

The models are trained on examples generated by the parallel training data. The labels for training the classifier of target word $t_j$ are defined as follows:
\begin{equation}
\label{BaselineLabel}
   label_{t_j}(s,t) =  \left\{
     \begin{array}{lr}
       1 & \text{if} \quad t_j \in t\\
       0 & \text{if} \quad t_j \notin t
     \end{array}
   \right.
\end{equation}
This model approximates the probability $p(t_j|s)$ of a target word  $t_j$ given the source sentence $s$. 
We will discuss our alternative method using neural network to estimate those probabilities in the next section.

In \cite{niehues2013mt}, the source context is considered in a way that the sentence is no longer represented by a bag of words, but by a bag of ngrams. 
Using this representation, they could integrate the order information of the words, but the dimension of the input space is increased.
We also adapt this extension to our model by encoding the bigrams and trigrams as ordinary words in the source vocabulary.

After inducing the probability for every word $t_j$ given the source sentence $s$, 
these probabilities were combined into the probability of the whole target sentence $t=t_1 \ldots t_J$ given $s$ as described in Section~\ref{sec:sentencelevel}.

\subsection{General network architecture}
\label{sec:NNDWLArchitecture}
After we reviewed the original DWL in the last section, we will now describe the neural network that replaces the MaxEnt model for calculating the probabilities $p(t_j|s)$.

\begin{figure}[h!]
\centering
\includegraphics[width=0.8\columnwidth]{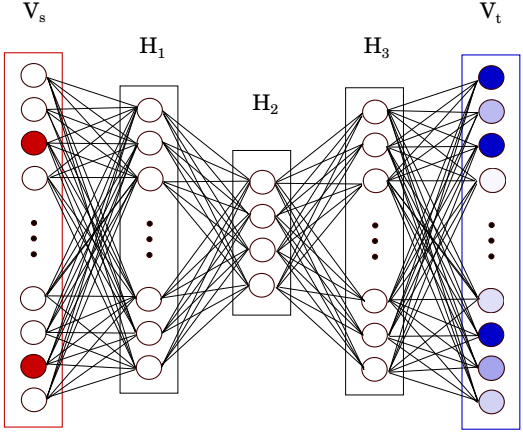}
\caption{\label{fig:FFNNArchitecture} {\it FFNN architecture for learning lexical translation.}}
\end{figure}
The input and output of our neural network-based DWL are the source and target sentences from which we would like to learn the lexical translation relationship.  
As in the original DWL approach, we represent each source sentence $s$ as a binary column vector $\vec{\hat{s}} \in \{0|1\}^{|V_s|}$ 
with ${V_s}$ being the considered vocabulary of the source corpus. 
If a source word $s_i$ appears in that sentence $s$, the value of the corresponding index $i$ in $\vec{\hat{s}}$ is $1$, and $0$ otherwise. 
Hence, the source sentence representation should be a sparse vector, depending on the considered vocabulary $V_s$. 
The same representation scheme is applied to the target sentence $t$ to get a sparse binary column vector $\vec{\hat{t}}$ with the considered target vocabulary ${V_t}$.

As the Figure~\ref{fig:FFNNArchitecture} depicts, our main neural network-based DWL architecture for learning lexical translation is a feed-forward neural network (FFNN) with three hidden layers. 
The matrix  $\vec{W}^{(1)} \in \mathbb{R}^{V_s\times|H_1|}$ connects the input layer to the first hidden layer. 
Two matrices $\vec{W}^{(2)} \in \mathbb{R}^{|H_1|\times|H_2|}$ and  $\vec{W}^{(3)} \in \mathbb{R}^{|H_2|\times|H_3|}$ encodes the learned translation mapping 
between two compact global feature spaces of the source and target contexts. And the matrix $\vec{W}^{(4)} \in \mathbb{R}^{|H_3|\times|V_t|}$ computes the lexical translation output. 
$|H_1|$, $|H_2|$, and $|H_3|$ are the number of units in the first, second and third hidden layers, respectively. 
The lexical translation distribution of the words in the target sentence $p(t_i|s)$ for a given source sentence $s$ is computed by a forward pass:
$$ p(t_i|s) = \sigma_i({\vec{W}^{(4)}}^T\vec{O}^{(3)}) $$
where:
\[
\begin{aligned}
&\vec{O}^{(k)} = \begin{bmatrix} \sigma_j ({\vec{W}^{(k)}}^T\vec{O}^{(k-1)}) \end{bmatrix}  \quad k \in \{1,2,3\} \\
&\vec{O}^{(0)} = \vec{\hat{s}} \quad \text{and} \quad\vec{O}^{(4)} = \vec{p}(t,s)
\end{aligned}
\] 
and $\sigma_j$ is the \textit{sigmoid} function $\sigma(x)$ applied to the $j^{th}$ value in a column vector:
$$ \sigma(x) = \frac{1}{1+\mathbf{e}^{-x}}$$

So the parameters of the network are:
$$ \theta = (\vec{W}^{(1)}, \vec{W}^{(2)}, \vec{W}^{(3)}, \vec{W}^{(4)})$$
 
To investigate the impact of the network configuration, we built a simpler architecture with only one hidden layer featuring the translation relationship between source and target sentences.  
We will refer this as the \textit{SimNNDWL} in the comparison section later.

\subsection{Network training}
In neural network training, for each instance, which is comprised of a sentence pair $(s,t)$, we maximize the similarity between the conditional probability $p_i = p_\theta(t_i|s)$ to either 1 or 0 
depending on the appearance of the corresponding word $t_i$ in the target sentence $t$. The neural network operates as a multivariate classifier 
which gives the probabilistic score for a binary decision of independent variables, i.e the appearances of target words.
Here we minimize the cross entropy error function between the binary target sentence vector $\vec{\hat{t}}$ and the output of the network $\vec{p}=\begin{bmatrix} p_i \end{bmatrix}$:
$$ E = -\frac{1}{V_t}\sum^{V_t}_{i=1}{(\vec{\hat{t}}_i \ln p_i + (1-\vec{\hat{t}}_i) \ln (1-p_i)})$$

We train the network by back-propagating the error based on the gradient descent principle. 
The error gradient for the weights between the last layer and the output is calculated as:

$$\frac{\partial E}{\partial w^{(4)}_{ij}} = (\vec{O}^{(4)}_j-\vec{\hat{t}}_j)\vec{O}^{(3)}_i$$

The error gradient for the weights between the other layers is calculated based on the error gradients for activation values from the previous layers:
$$\frac{\partial E}{\partial w^{(k)}_{ij}} = \frac{\partial E}{\partial \vec{O}^{(k)}_{j}}\vec{O}^{(k-1)}_{i} $$

Then the weight matrices are batch-updated after each epoch: 
$$\vec{W}^{(k)}[T+1] = \vec{W}^{(k)}[T] - \eta\sum^N_{i=1} \frac{\partial E}{\partial \vec{W}^{(k)}}$$
where:
\begin{itemize}
  \item $N$ is the number of training instances.
  \item $\eta$ is the learning rate of the network.
  \item $\vec{W}^{(k)}[T+1]$ is the weight matrix of the layer $k$ after $T+1$ epochs of training.
\end{itemize}

\subsection{Sentence-level lexical translation scoring}
\label{sec:sentencelevel}
With the independence assumption among target words, the target probabilities are combined to form the sentence-level lexical translation score:
\begin{equation} \label{eq:1}
p(t|s) = \prod_{t_j \in v_t}  p(t_j|s)
\end{equation}
where $v_t$ is the set of all target words appearing in the target sentence $t$.

In Equation~\ref{eq:1}, we need to update the lexical translation score only if a new word appears in the hypothesis. 
That means we do not take into account the frequency of words but multiply the probability of one word only once even if the word occurs several times in the sentence. 
Other models in our translation system, however, will restrict overusing a particular word. 
Furthermore, to keep track of which words whose probabilities have been calculated already, additional book keeping would be required. 
In order to avoid those difficulties, we come up with the following approximation given $J$ is the length of the target sentence $t$:
\begin{equation} \label{eq:2}
p(t|s) = \prod^J_{j=1} p(t_j|s)
\end{equation}

In order to speed up the calculation of the target word probabilities, we pre-calculate all probabilities for a given source sentence prior to translations. 
In a naive approach we would need to pre-calculate the probabilities for all possible target words given the source sentence. This would lead to a very slow calculations. 
Therefore, we first define the target vocabulary of a source sentence as the vocabulary comprised of the respective words from the phrase pairs matching to the source sentence.
Using this definition, we only need to pre-calculate the probabilities of all words in the target side of the phrase table and not all target words in the whole corpus.
And we can calculate the score for every phrase pair even before starting with the translation.

\section{Experiments}
\label{sec:experiments}
In this section, we describe the translation system we use for the experiments, the configurations of the NNDWL and the results of those experiments.
\subsection{System description}
The system we use as our baseline is a state-of-the-art translation system for English to French without any DWL. 
To the baseline system, we add several DWL components trained on different corpora as independent features in the log-linear framework utilized by our in-house phrase-based decoder.

The system is trained on the EPPS, NC, Common Crawl, Giga corpora and TED talks\cite{TED2012}. The monolingual data we used to train language models includes the corresponding monolingual parts
of those parallel corpora plus News Shuffle and Gigaword. The data is preprocessed and the phrase table is built using the scripts from the Moses package \cite{ref:MosesDecoder}.  
We adapt the general, big corpora to the in-domain TED data using the Backoff approach described in \cite{niehues2012}. 
Adaptation is also conducted for the monolingual data. We train a 4-gram language model using the SRILM toolkit \cite{ref:srilm}. 
In addition, several non-word language models are included to capture the dependencies between source and target words and reduce the impact of data sparsity. 
We use a bilingual language model as described in \cite{niehues2011} as well as a cluster language model based on word classes generated by the MKCLS algorithm \cite{mkcls}.
Short-range reordering is performed as a preprocessing step as described in \cite{ref:rottmann2007}. 

Our in-house phrase-based decoder is used to search for the best solutions among translation hypotheses and the optimization of the 13 to 17 features, depending on the settings we use,
is performed using Minimum Error Rate Training \cite{ref:och03mer}. 
The weights are optimized and tested on two separate sets of TED talks. The development set consists of 903 sentences containing 20k words. 
The test set consists of 1686 sentences containing 33k words.

We investigate the impact of our approach by employing different configurations of the neural networks described in details in the following section. 
We then evaluate those configurations not only for English$\rightarrow$French but also for English$\rightarrow$Chinese and German$\rightarrow$English 
with similar translation system setups. 

Our NNDWL models are trained on a small subset of the mentioned training corpora, mainly the TED data. 
Although the TED corpus is quite small compared to the overall training data, it is very important since it matches best the test data. 
In order to speed up the process of testing different configurations, we therefore train the NNDWL only on this corpus except for the comparison reported in Section~\ref{sec:data}.
The statistics of the training and validation data for the NNDWL are shown in Table~\ref{table:statistics}.\\

\begin{table} [h!] 
\label{table:statistics}
\centerline{ 
\begin{tabular}{cc|c|c|c|}
\cline{3-5}
\cline{3-5}
& & \textbf{En-Fr} & \textbf{En-Zh} & \textbf{De-En}  \\ 
\cline{1-5}
\multicolumn{1}{ |c  }{\multirow{2}{*}{\textbf{Training}} } &
\multicolumn{1}{ |c| }{Sent.} & 149991 & 140006 & 130654   \\ 
\cline{2-5}
\multicolumn{1}{ |c  }{}                        &
\multicolumn{1}{ |c| }{Tok. (avg.)} & 3.1m & 3.3m & 2.5m   \\ 
\cline{1-5}
\cline{1-5}
\multicolumn{1}{ |c  }{\multirow{2}{*}{\textbf{Validation}} } &
\multicolumn{1}{ |c| }{Sent.} & 6153 & 8962 & 7430   \\ 
\cline{2-5}
\multicolumn{1}{ |c  }{}                        &
\multicolumn{1}{ |c| }{Tok. (avg.)} & 125k & 211k & 142k   \\ 
\cline{1-5}
\end{tabular}}
\caption{\label{table:statistics} {\it Statistics of the corpora used to train NNDWL}}
\end{table}

\subsection{Network configurations}
In our main neural network architecture we proposed, the sizes of the hidden layers $|H_1|$, $|H_2|$, $|H_3|$ are $1000$, $500$, $1000$, respectively. 
If we use the original source and target vocabularies, for the English$\rightarrow$French direction trained on preprocessed TED 2013 data, $V_s$ includes $47957$ words and $V_t$ includes $62660$ words.
Because of the non-linearity calculations through such a large network, the training is extremely time-consuming. 
In order to boost the efficiency, we limit the source and target vocabularies to the most frequent ones. All words outside the lists are treated as unknown words. 
We vary the size of the considered vocabularies from the values $\{500, 1000, 2000, 5000\}$ 
while keeping the sizes of the hidden layers the same (i.e. $1000\times500\times1000$). 
In preliminary experiments, this layout lead to the best performance. So we used this layout for the remaining of the paper.

The same calculation problem occurs with the source contexts, even more seriously due to the curse of dimentionality.
Hence, we applied the same cut-off scheme to the source-side bigrams and trigrams with the most-frequent bigram and trigram numbers set at $(200,100)$, $(500,200)$ and $(1000,500)$.

The simpler architecture \textit{SimNNDWL} consisting of one 1000-unit hidden layer is compared to the main architecture with the same setup.

For training our proposed architecture, the gradient descent with a batch size of $15$ and a learning rate of $0.02$ is used. 
Gradients are calculated by averaging across a minibatch of training instances and the process is performed for 35 epochs. 
After each epoch, the current neural network model is evaluated on a separate validation set, 
and the model with the best performance on this set is utilized for calculating lexical translation scores afterwards. 
We regularize the models with the $L_2$ regularizer. As an alternative to the $L_2$, we also experiment with the dropout technique \cite{hinton2012}, 
where the neurons in the last hidden layer are randomly dropped out with the probability of $0.4$. However, it did not help as indicated by its performance on the system later.
The training is done on GPUs using the Theano Toolkit\cite{bergstra:2010}.

\subsection{Results}
Here we report the results using different NNDWL configurations mainly for an English$\rightarrow$French translation system. 
We also report the results using the best configurations for other language pairs. 
\subsubsection{Experiments with different vocabulary sizes}
The results of the English$\rightarrow$French translation system with NNDWL models trained with different vocabulary sizes are shown in Table~\ref{table:vocabs}.

\begin{table} [h!] 
\label{table:vocabs}
\centerline{ 
\begin{tabular}{l|cc} \hline
System (En-Fr) & BLEU & $\Delta$BLEU \\ \hline \hline 
\textit{Baseline} & 31.94 & -- \\ \hline  
MaxEnt DWL & 32.17 & +0.23 \\ \hline
NNDWL 500 & 32.06 & +0.12 \\
NNDWL 1000 & 32.37 & +0.43 \\
NNDWL 2000 & \textbf{32.38} & \textbf{+0.44} \\
NNDWL 5000 & 32.07 & +0.13 \\ \hline
Full NNDWL & 32.06 & +0.12 \\ \hline
\end{tabular}}
\caption{\label{table:vocabs}{\it Results of the English$\rightarrow$French NNDWL.}}
\end{table}
Varying the vocabulary sizes for both source and target sentences not only helps to dramatically reduce neural network training time but also affects the translation quality.
In our experiments, neural networks with 1000- and 2000-most-frequent-word vocabularies show the biggest improvements with around 0.44 BLEU points in translating from English to French. 
They perform better than the DWL using the maximum entropy approach and the NNDWL with the whole source and target vocabularies. 

While all NNDWL models achieve notable BLEU gains compared to the strong baseline, some of them are worse than the original MaxEnt model. 
It might be due to the fact that the original MaxEnt model uses the source contexts whereas the NNDWL models uses just the source words. 
\subsubsection{The impact of $n$-gram source contexts}
Tables~\ref{table:sc2000} and~\ref{table:sc1000} show the impact of bigrams and trigrams extracted from source sentences. 
We also vary the numbers of the bigrams and trigrams which appeared most often.

\begin{table} [h!] 
\label{table:sc2000}
\centerline{ 
\begin{tabular}{l|cc} \hline
System (En-Fr) & BLEU & $\Delta$BLEU \\ \hline \hline 
\textit{Baseline} & 31.94 & --  \\  \hline
NNDWL 2000 & \textbf{32.38} & \textbf{+0.44} \\ \hline
NNDWL 2000 SC-200-100 & 32.35 & +0.41 \\
NNDWL 2000 SC-500-200 & \textbf{32.44} & \textbf{+0.50} \\
NNDWL 2000 SC-1000-500 & 32.36 & +0.42 \\ \hline
\end{tabular}}
\caption{\label{table:sc2000} {\it Results of the 2000-NNDWL with source contexts.}}
\end{table}
For the NNDWL model with 2000-most-frequent-word vocabularies, including source contexts helps in some cases and does not harm the translation performance in the other cases. 
With the 500 most-frequent bigrams and 200 most-frequent trigrams, we achieve the best improvements of $0.5$ BLEU points over the baseline.

\begin{table} [h!] 
\label{table:sc1000}
\centerline{ 
\begin{tabular}{l|cc} \hline
System (En-Fr) & BLEU & $\Delta$BLEU \\ \hline \hline 
\textit{Baseline} & 31.94 & -- \\  \hline
NNDWL 1000 & \textbf{32.37} & \textbf{+0.43} \\ \hline
NNDWL 1000 SC-200-100 & 32.01 & +0.07  \\
NNDWL 1000 SC-500-200 & 32.23 & +0.29 \\
NNDWL 1000 SC-1000-500 & \textbf{32.39} & \textbf{+0.45} \\ \hline
\end{tabular}}
\caption{\label{table:sc1000} {\it Results of the 1000-NNDWL with source contexts.}}
\end{table}
The gains from adding source contexts to the 1000-vocabulary-size NNDWL model are not clearly observed as in the case of the 2000-vocabulary-size model. 
This might indicate that we should set the numbers of the source contexts to be proportional somehow with the size of the vocabularies.

\subsubsection{The impact of using different architectures}
\begin{table} [h!] 
\label{table:architecture}
\centering
\begin{tabular}{l|cc} \hline
System (En-Fr) & BLEU & $\Delta$BLEU  \\ \hline \hline 
\textit{Baseline} & 31.94 & --\\ \hline
NNDWL 1000 & \textbf{32.37} & \textbf{+0.43} \\
SimNNDWL 1000 & 32.12 & +0.18 \\ \hline
NNDWL 2000 & \textbf{32.38} & \textbf{+0.44}\\
SimNNDWL 2000 & 32.29 & +0.35\\ \hline
NNDWL 5000 & \textbf{32.07} & \textbf{+0.13}\\ 
SimNNDWL 5000 & 31.71 & -0.23 \\ \hline
\end{tabular}
\caption{\label{table:architecture} {\it Results of NNDWL and SimNNDWL architectures.}}
\end{table}

Here we compare our main architecture with the simpler architecture \textit{SimNNDWL} consisting of one 1000-unit hidden layer. 
While the \textit{SimNNDWL} trains faster (157 hours vs. 202 hours for training English$\rightarrow$French with the whole vocabularies), translation time performance is not significantly affected. 
Since there are decreases in BLEU score using \textit{SimNNDWL} architecture as shown in Table~\ref{table:architecture}, the deep architecture seems to have an advantage over the simple architecture.
Hence, we stick with our main architecture for remaining experiments.

\subsubsection{The impact of data used to train NNDWL models}
\label{sec:data}
We also train our NNDWL models on a bigger corpus concatinating EPPS, NC and TED. The results in Table~\ref{table:bigdata} shows that using a bigger corpus does not improve the translation quality. 
The DWL models trained on in-domain data only, i.e. TED, perform similar or better than the models trained on more data but broader domains. 
This observation also holds true for original the \textit{MaxEnt DWL} models reported in \cite{ha2013kit}. 

\begin{table} [h!] 
\centerline{ 
\begin{tabular}{l|cc} \hline
System (En-Fr) & BLEU & $\Delta$BLEU \\ \hline \hline 
\textit{Baseline} & 31.94 & -- \\  \hline
NNDWL 1000 on TED & \textbf{32.37} & \textbf{+0.43} \\
NNDWL 1000 on EPPS+NC+TED & 32.33 & +0.39\\ \hline
\end{tabular}}
\caption{\label{table:bigdata} {\it Results of the NNDWL trained on different corpora.}}
\end{table}

\subsubsection{Other language pairs}
We conducted the experiments with NNDWL models mainly on our English-to-French translation system in order to investigate the impact of our method on a strong baseline. 
However, we would like to inspect the effect of the DWL on language pairs with long-range dependencies or differences in word order. 

For that purpose, we built similar NNDWL models and integrate them to our translation systems for other language pairs. 
Tables~\ref{table:EnZh} and~\ref{table:DeEn} show the results of English$\rightarrow$Chinese and German$\rightarrow$English, respectively. \\

\textbf{English$\rightarrow$Chinese}
\begin{table} [h!] 
\centerline{ 
\begin{tabular}{l|cc} \hline
System (En-Zh) & BLEU & $\Delta$BLEU \\ \hline \hline 
Baseline & 17.18 & -- \\  \hline
MaxEnt DWL & 16.78 & -0.40 \\ \hline
NNDWL 500 & 17.09 & -0.09 \\
NNDWL 1000 & 17.58 & +0.40 \\
NNDWL 1000 SC-200-100  & \textbf{17.63} & \textbf{+0.45}  \\
NNDWL 2000 & 17.26 & +0.08 \\
NNDWL 2000 SC-200-100 & 17.20 & +0.02 \\ \hline
\end{tabular}}
\caption{\label{table:EnZh} {\it Results of the English$\rightarrow$Chinese NNDWL}}
\end{table}

In case of the English$\rightarrow$Chinese direction, the NNDWL significantly improves the translation quality, with an increment of 0.45 BLEU points over the baseline. 
That best BLEU gain comes from the NNDWL with 1000-most-frequent-word vocabularies and the source contexts containing 200 bigrams and 100 trigrams. \\

\textbf{German$\rightarrow$English} \\

In case of the German$\rightarrow$English direction, the NNDWL also helps to gain 0.34 BLEU points over the baseline 
with the best model (i.e. 2000 most-frequent-word vocabularies with source contexts). 
However, the improvements is not notably different compared to the original MaxEnt DWL. \\

\begin{table} [h!] 
\centerline{ 
\begin{tabular}{l|cc} \hline
System (De-En) & BLEU & $\Delta$BLEU \\ \hline \hline 
Baseline & 29.70 & -- \\  \hline
MaxEnt DWL & 29.95 & +0.25 \\ \hline
NNDWL 500 & 29.82 & +0.12 \\
NNDWL 1000 & 29.92 & +0.22  \\
NNDWL 2000 & 29.95 & +0.25 \\
NNDWL 2000 SC-500-200 & \textbf{30.04} & \textbf{+0.34} \\
NNDWL 5000 & 29.89 & +0.19 \\ \hline
\end{tabular}}
\caption{\label{table:DeEn} {\it Results of the German$\rightarrow$English NNDWL}}
\end{table}

\section{Conclusion}
\label{sec:conclusion}
In this paper we described a deep neural network approach for DWL modeling and the integration into a standard phrase-based translation system.
Using neural networks as a non-linear classifier for DWL enables the ability of learning the abstract representation of global contexts and their dependencies. 
We investigated various network configurations on different language pairs. 
When we deployed our best NNDWL model as a feature in our decoder, it helps to improve up to 0.5 BLEU points compared to a very strong baseline. 

Our NNDWL does not require linguistic resources nor feature engineering. Thus, it can easily be ported to new languages. 
Furthermore, the probability calculation can be done in a preprocessing step. Therefore, the new model would not significantly slow down the translation process.
Although we do not feature linguistic resources in our NNDWL, they can be useful in modeling the translation probability of the languages from which they are avalaible. 
In future work we will try to integrate linguistic features into the model. Moreover, context vector of words might be helpful in further reducing the data sparseness problem.

\section{Acknowledgments}
The research leading to these results has received funding from the European Union Seventh Framework Programme (FP7/2007-2013) under grant agreement n$^\circ$ 287658.

\bibliographystyle{IEEEtran}
\bibliography{references}

\end{document}